# MR-GNF: Multi-Resolution Graph Neural Forecasting on Ellipsoidal Meshes for Efficient Regional Weather Prediction

**Andrii Shchur[1], Inna Skarga-Bandurova[2]**

[1]Kyiv School of Economics

[2]Oxford Brookes University

ashchur@kse.org.ua, iskarga-bandurova@brookes.ac.uk

**Abstract**

Weather forecasting offers an ideal testbed for artificial intelligence (AI) to learn complex, multi-scale physical systems. Traditional numerical weather prediction remains computationally costly for frequent regional updates, as high-resolution nests require intensive boundary coupling. We introduce Multi-Resolution Graph Neural Forecasting (MR-GNF), a lightweight, physics-aware model that performs short-term regional forecasts directly on an ellipsoidal, multi-scale graph of the Earth. The framework couples a 0.25° region of interest with a 0.5° context belt and 1.0° outer domain, enabling continuous cross-scale message passing without explicit nested boundaries. Its axial graph-attention network alternates vertical self-attention across pressure levels with horizontal graph attention across surface nodes, capturing implicit 3-D structure in just 1.6 M parameters. Trained on 40 years of ERA5 reanalysis (1980-2024), MR-GNF delivers stable +6 h to +24 h forecasts for near-surface temperature, wind, and precipitation over the UK-Ireland sector. Despite a total compute cost below 80 GPU-hours on a single RTX 6000 Ada, the model matches or exceeds heavier regional AI systems while preserving physical consistency across scales. These results demonstrate that graph-based neural operators can achieve trustworthy, high-resolution weather prediction at a fraction of NWP cost, opening a practical path toward AI-driven early-warning and renewable-energy forecasting systems.

**Code** — https://github.com/AndriiShchur/MR-GNF

## Introduction

Artificial intelligence (AI) is rapidly transforming weather prediction by learning to emulate atmospheric dynamics at a fraction of the computational cost of traditional numerical models. However, producing reliable short-term regional forecasting remains a significant challenge (Allen et al., 2025; Zhang et al., 2025). Operational centres can deliver impressive global skill using physics-based numerical weather prediction (NWP), but running convection-aware resolution everywhere is computationally expensive and limits the frequency of updates. In practice, regional stakeholders, such as energy and emergency management agencies, require forecasts that combine fine spatial detail with rapid refresh cycles over specific areas of interest, without the full cost of global high-resolution modelling. Classical limited-area NWP addresses this need by nesting high-resolution domains within coarser global models. However, this approach carries two persistent drawbacks: (i) computational cost scales steeply with inner-domain resolution and update frequency, and (ii) lateral boundary conditions can introduce artefacts or damp physically important waves, particularly in complex coastal and orographic regions or during fast-evolving synoptic events (Davies, 1976; Warner et al., 1997; Termonia et al., 2009; Marsigli et al., 2014; Yano et al., 2018; Zhang et al., 2024).

Recent advances in AI provide a data-driven alternative to traditional numerical weather prediction. Foundation-scale weather models such as GraphCast (Lam et al. 2023), EWMoE (Gan et al., 2024), GenCast (Price et al. 2025), and Pangu-Weather (Bi et al. 2023) achieve state-of-the-art global skill with orders-of-magnitude faster runtimes than NWP. Architectures based on adaptive Fourier operators such as FourCastNet (Pathak et al., 2022) further underscore this trend toward learned, global-scale surrogates. Although these models are optimised for medium-range, planet-wide inference, they remain too heavy for high-frequency, short-range regional updates, where retraining speed, localised precision, and efficient deployment are critical.

A complementary thread explores non-uniform and graph-based discretizations (Nipen et al., 2024) that focus resolution where it matters most, pointing to a new paradigm for adaptive, data-driven regional forecasting. Graph neural networks (GNNs) naturally represent the irregular geometry of the Earth's surface and can support adaptive resolution, but most existing designs operate at a single scale, neglecting the hierarchical structure of atmospheric dynamics. This motivates the development of multi-resolution, geometry-aware architectures for regional forecasting.

In this work, we present Multi-Resolution Graph Neural Forecasting (MR-GNF), a lightweight framework for short-term regional weather prediction built directly on an ellipsoidal mesh of the Earth. The proposed approach advances scientific understanding of how adaptive, geometry-aware neural architectures can emulate complex geophysical dynamics, offering a scalable foundation for future regional climate and environmental modelling.

The tri-band ellipsoidal mesh couples a 0.25° ROI with a 0.5° context belt (±6° lon, ±4° lat) and a 1.0° outer domain, generated via spacing-field optimisation using JIGSAW. This design preserves geodesic continuity and enables cross-scale message passing without explicit nested boundaries.

At the model level, an axial graph-attention network alternates vertical self-attention (across pressure-level tokens) and horizontal graph attention (across surface nodes), achieving implicit 3D coupling in a computationally compact form. The system was trained on 40 years of ERA5 reanalysis (Hersbach et al., 2020) using six surface and fifteen pressure-level variables, augmented with orography-aware static features from ETOPO (elevation, slope, relief, and a refined land–water mask) and 2D positional encodings. The model is tailored to short-term regional forecasting over a Britain-centric Euro-Atlantic ROI, performing +6 h one-step predictions and +24 h autoregressive rollouts (K = 4). Performance is evaluated using Root Mean Square Error (RMSE) and Mean Absolute Error (MAE) across near-surface and upper-air fields.

This paper makes the following key contributions:

- We propose a tri-band ellipsoidal mesh design generated via spacing-field optimisation, which enables boundary-free cross-scale coupling through graph connectivity.
- We design a lightweight axial graph-attention core that alternates vertical self-attention with horizontal graph attention, achieving implicit 3D coupling while remaining computationally compact.
- The MR-GNF provides cross-scale forecasting without cascades. The system performs +6 h one-step prediction and autoregressive rollouts to +24 h, exchanging information between scales directly through graph connectivity rather than cascaded global-regional stages.
- With ~1.6 M parameters and < 80 GPU-hours total training, the model achieves 6-24 h skill comparable to heavy regional AI systems, demonstrating a practical path toward low-latency, high-resolution forecasting for storm early-warning and wind-energy operations.

## Related Work

**Global AI weather models**
Data-driven global forecasters have advanced rapidly in both skill and efficiency. GraphCast delivers state-of-the-art medium-range skill at 0.25° resolution while running orders of magnitude faster than full NWP (Lam et al., 2023). Pangu-Weather extends performance further through 3-D neural networks that explicitly capture vertical dependencies (Bi et al. 2023). AFNO-based FourCastNet pioneered fast global inference at 0.25° (Pathak et al. 2022), while foundation-style approaches such as ClimaX(Nguyen et al. 2023) explore flexible pretraining and adaptation across tasks (Nguyen et al. 2023). Aurora (Bodnar et al., 2024) extends this "foundation model" line by training a 3D architecture on over a million hours of heterogeneous Earth-system data. It matches or surpasses operational forecasts for high-resolution weather, air quality and ocean waves while running at much lower computational cost. Operational centres are also building ML systems, e.g., ECMWF AIFS (Lang et al. 2024).

**Probabilistic and diffusion ensembles**
Beyond deterministic forecasting, probabilistic generative models have shown growing maturity. GenCast (Price et al. 2025) applies a diffusion-based probabilistic model that exceeds the ECMWF ENS on many medium-range metrics while retaining fast runtimes. Related diffusion-ensemble work (Andrae et al. 2025) underscores the growing maturity of generative ML for weather uncertainty. However, these systems are typically global and heavyweight, and tailoring them for frequently updated, high-resolution regional products can still imply cascaded procedures or full-domain inference.

**Regional, non-uniform, and graph-based modelling**
A complementary direction targets regional skill using non-uniform grids and graph neural networks (GNN). Nipen et al. (2024) propose a stretched-grid GNN that concentrates resolution over a Nordic ROI while preserving coarser context elsewhere, demonstrating strong regional skill and operational realism. Learning on multi-scale geospatial graphs, therefore, represents a crucial next frontier, enabling spatial interactions to be modelled consistently across resolutions. while maintaining coarser context elsewhere, reporting strong regional performance and operationally relevant behaviour – evidence that graph-based, multi-resolution designs can bridge local detail with large-scale flow (Nipen et al. 2024).

**Our position.** We target the niche between global foundation models (e.g., GraphCast, GenCast) and classical nested regional NWP. Similar to stretched-grid regional AI (e.g., Nipen et al. 2024), we focus resolution over a ROI, but extend this concept by: (i) employing an explicit three-band ellipsoidal mesh (0.25° ROI, 0.5° context ±6°/±4°, 1.0° outer domain); (ii) enabling cross-scale message passing instead of lateral boundary conditions, promoting stable ROI-context exchange; (iii) keeping the model operationally lightweight (~1.6 M params) and single-stage (no cascaded fine-tuning), trained for +6 h and evaluated via short AR rollouts (6-24 h) to suit frequent regional updates; and (iv) integrating orography-aware static features and tailored 2-D positional encodings to preserve coastal/orographic structure. In short, compared to global heavy models, MR-GNF trades the medium-range reach of large foundation models for operationally cheap, high-resolution regional skill, and compared to nested NWP, it avoids costly inner-domain time stepping and boundary artefacts by using a multi-resolution graph that natively couples local detail with large-scale flow.

## Preliminaries

This section details the datasets, spatial representation, and experimental setup used to train and evaluate MR-GNF for short-term regional weather forecasting.

**Dataset, variables, and targets**

The study focuses on a UK-centric ROI (49-59° N, 11° W-2° E). We use the ERA5 reanalysis dataset (Hersbach et al., 2020) produced by the European Centre for Medium-Range Weather Forecasts (ECMWF) as the ground truth for model training and evaluation. ERA5 provides global atmospheric fields at 0.25° spatial resolution (721 × 1440 latitude-longitude grid points) and 6-hourly cadence.

Inputs and targets are extracted from ERA5 and mapped to the tri-band mesh. Each sample include six surface variables: 2 m temperature (t2m), 2 m dewpoint (d2m), mean sea-level pressure (msl), 10 m wind components (u10, v10), and logarithmic total precipitation (tp_log); 15 pressure-level variables: temperature (t), wind components (u, v), relative humidity (r), and geopotential (z) at 850 hPa, 500 hPa, and 300 hPa. Parameter tp_log is used to stabilise the heavy-tailed precipitation skew while retaining dry events.

Temporal splits: 1980-2018 (training), 2019-2023 (validation), and 2024 (test).

**Forecasting objective and model inputs**

The objective is to predict the short-term atmospheric state over the ROI up to 24 hours ahead, using recent spatiotemporal observations.

Let the ROI domain be $\Omega_{ROI} \subset S^2$, embedded within the tri-band mesh $\Omega = \Omega_{ROI} \cup \Omega_{belt} \cup \Omega_{outer}$. At time t, the multivariate atmospheric state is denoted $X_t \in \mathbb{R}^{C \times N}$, where C is the number of variables and N the number of mesh nodes. Given a short input history $H_t = \{X_{t-T_{in}+1}, \ldots, X_t\}$, the forecasting model $f_\theta$ aims to predict the next $T_{out}$ future states $\hat{Y}_t = \{\hat{X}_{t+1}, \ldots, \hat{X}_{t+T_{out}}\}$, with $T_{in}$ = 2 (12 h) and $T_{out}$ = 1 (+6 h).

The model parameters $\theta$ are optimised by minimising the expected forecasting loss

$$\mathcal{L}(\theta) = \frac{1}{T_{out}} \sum_{k=1}^{T_{out}} \ell(\hat{X}_{t+k}(\theta), X_{t+k}),$$

where $\ell(\cdot)$ denotes a per-channel MSE for the core and its task-specific adaptations for the specialised heads.

During evaluation, predictions are autoregressively rolled out to a total horizon of $T_{eval}$ = 4 (+24 h) using $\hat{X}_{t+k+1} = f_\theta(\hat{X}_{t+k})$, i.e., each forecast step is fed back as input for the next. Model skill is assessed using RMSE and MAE across all surface and upper-air channels over $\Omega_{ROI}$.

**Static features and positional encodings**

To incorporate coastal and orographic structure, we include four static channels (standardized elevation, slope, and relief from ETOPO1 (Amante & Eakins, 2009), and refined land-water mask derived from Natural Earth oceans/lakes with 2 km buffered river centerlines) and four 2D positional encodings (PEs): $\{\sin \varphi, \cos \varphi, \sin(\lambda \cdot \cos \bar{\varphi}), \cos(\lambda \cdot \cos \bar{\varphi})\}$, where $\lambda$ and $\varphi$ denote longitude and latitude respectively, and domain-mean latitude defined as $\bar{\varphi}$.

Statics and PEs are prepared at 1.0°/0.5°/0.25° and sampled to nodes. Latitude/longitude grids and provenance metadata are stored alongside.

**Mesh quality and ERA5-to-graph mapping**

The spatial domain is discretised using a tri-band ellipsoidal mesh comprising fine (0.25°), intermediate (0.5°), and coarse (1.0°) resolution bands. The experimental mesh contains N = 4,227 surface nodes and F = 8,240 triangles (~12.4 k undirected edges) with a mean degree of 5.85. Mesh quality is high: mean min/max angles 55.4°/65.1°, area-edge compactness $Q_{area/len} \approx 0.988$ (1.0 ideal), and mean edge-length-to-target ratio $h_r \approx 1.01$, indicating close adherence to the prescribed spacing field and minimal anisotropy.

ERA5 fields at 0.25°, 0.5°, and 1.0° resolution are bilinearly sampled to the corresponding mesh nodes by zone (ROI/belt/outer). A refined land-water mask reduces coastal bleed-through. Each variable is standardised using a streaming Welford estimator on the training window (Welford, 1962; Chan et al., 1983). The resulting feature tensors are shaped as $[T_t, C, N]$ and batched as $[B, T_t, C, N]$ for direct input into the graph model. Evaluation metrics (RMSE and MAE) are computed at +6, +12, +18, and +24 h across all surface and upper-air channels over the ROI.

## Methods

The pipeline comprises: (i) construction of a multi-resolution ellipsoidal mesh; (ii) zone-aware sampling that maps ERA5 and static layers to nodes while preserving native resolution and coastal sharpness; (iii) a lightweight spatio-vertical graph neural core with task heads for wind and precipitation, trained for +6 h and rolled out to +24 h.

Unlike uniform global meshes or nested regional domains, this approach embeds adaptive spatial resolution directly into graph topology, enabling the GNN to exchange information between coarse and fine scales through connectivity rather than through explicit lateral boundary conditions, and form the structural basis for cross-scale message passing in later stages.

**Tri-band ellipsoidal mesh**

We define a multi-resolution triangulated mesh on the Earth's ellipsoid that concentrates resolution over the ROI while remaining computationally light elsewhere.

ROI bounds are defined as $B = [\lambda_{min}, \lambda_{max}] \times [\varphi_{min}, \varphi_{max}]$, with context margins of (±6° lon; ±4° lat) on an ellipsoid with semi-axes $(a, b)$. We specify a target spacing field $s(\phi, \lambda)$ that controls node density across three zones:

$$s(\phi, \lambda) = \begin{cases} 0.25°, & (\phi, \lambda) \in B, \\ 0.5°, & (\phi, \lambda) \in B \oplus (\pm 6°, \pm 4°) \setminus B, \\ 1.0°, & \text{otherwise.} \end{cases}$$

This defines the tri-band hierarchy of a 0.25° ROI, a 0.5° context belt, and a 1.0° outer domain. A Delaunay-quality triangulation is generated with JIGSAW (Engwirda, 2016) using the target spacing field. Geographic coordinates $(\varphi, \lambda)$ are mapped to the reference ellipsoid and the mesh is cropped to $B$ plus its context margins. Each node is tagged with a zone label Z ∈ {ROI, belt, outer}. The resulting mesh is represented as M = (V, E, F, Z), where V, E, and F denote vertices, edges, and triangular faces, respectively.

Mesh quality is verified using standard triangle metrics on the sphere: min/max angle distributions, area-edge-length compactness, relative edge length vs. target spacing, and vertex degree. In our build, the mesh achieves near-equilateral quality in the ROI/belt (typical min angle ~55°, 5-95% between ~47-60°, compactness ≈0.99) and stable node degree (~5-6), indicating low numerical anisotropy and well-controlled adaptation across resolution transitions. Figure 1 illustrates the resulting tri-band ellipsoidal mesh used in all experiments.

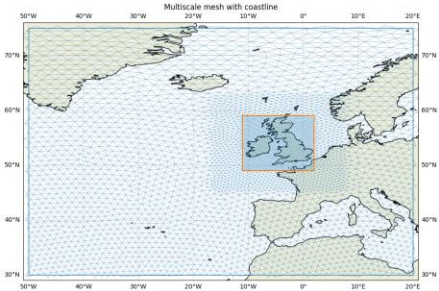

Figure 1: Multi-resolution ellipsoidal mesh (UK-centric ROI).

**Zone-aware sampling**
To preserve native resolution and avoid cross-scale aliasing, we transfer both meteorological fields (ERA5) and static layers (orography and positional encodings) from rectilinear latitude-longitude grids to the irregular tri-band mesh (Fig. 2) using a zone-aware bilinear sampler. The sampler performs bilinear interpolation per channel, applies coast-aware masking, and outputs a unified node tensor [$T_t$, C, N] for the GNN; with batch size B, inputs to the model have shape [B, $T_t$, C, N].

Let $\mathcal{G}_s$ denote a rectilinear lat-lon grid at resolution $\Delta x_s \in \{1.0°, 0.5°, 0.25°\}$ with monotonically increasing axes $\Lambda_s = \{\lambda_j\}$ (longitudes wrapped to $[-180,180)$) and $\Phi_s = \{\phi_i\}$. The mesh contains surface vertices $\{n\}_{n=1}^N$ with coordinates $(\lambda_n, \phi_n)$ and a zone label $z(n) \in \{\text{outer}, \text{belt}, \text{ROI}\}$. We associate each zone with its native grid as:

$$\mathcal{G}_\text{outer} \to 1.0°, \mathcal{G}_\text{belt} \to 0.5°, \mathcal{G}_\text{ROI} \to 0.25°.$$

For any scalar field $f$ tabulated on $\mathcal{G}_s$, the value at the node $n$ is obtained by bilinear interpolation on the zone-appropriate grid:

$$\tilde{f}(n) = (1 - t_x)(1 - t_y)f_{i_0,j_0} + t_x(1 - t_y)f_{i_0,j_1} + (1 - t_x)t_y f_{i_1,j_0} + t_x t_y f_{i_1,j_1},$$

where $t_x$ and $t_y$ are normalised horizontal offsets between grid points.

Vectors (e.g., $u_{10}$, $v_{10}$) are interpolated per component in their native basis. Pressure-level variables $g \in \{t, u, v, r, z\}$ use nearest-level selection $\ell^* \in \{850, 500, 300 \text{ hPa}\}$ followed by the same horizontal interpolation applied to $g_\ell$. For precipitation, we first clip numerical negatives and work in the stabilised log space $\text{tp\_log} = \log(1 + \max(\text{tp}, 0))$.

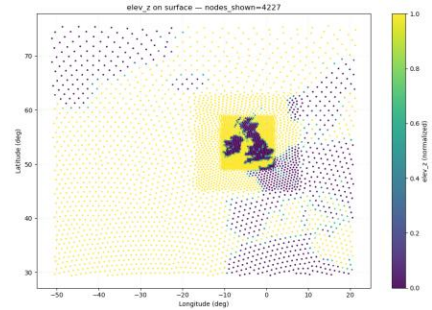

Figure 2: Zone-aware ERA5-to-graph sampler. Nodes in each zone pull from their native grids (ROI: 0.25°, belt: 0.5°, outer: 1.0°); coast-aware interpolation preserves sharp boundaries.

**Unified spatio-vertical graph neural core**
Each forecast is produced by a compact axial graph-attention network that alternates vertical and horizontal mixing.
**Input staging and static injection.** We concatenate a short input history ($T_\text{in}$) along the channel axis and project it to an embedding of width E using a $1 \times 1$ pointwise conv (implemented as a small MLP over features on the graph). Static features are encoded by a shallow Conv–Norm–ReLU tower and broadcast-added to the dynamic stream at every level. This design keeps the model resolution, while allowing orographic and positional structure to act as persistent physical biases that guide learning.
**Spatio-vertical coupling via axial graph blocks.** The network core consists of $L$ Graph-Axial GAT blocks. Each block alternates between two complementary operations:
- Vertical self-attention across the pressure-level tokens at each node (mixing the column) and
- Horizontal graph attention across horizontal neighbours within each level (mixing the flow on the mesh), followed by a lightweight feed-forward (3D-Conv/MLP) with residual connections and normalisation.

Residual feed-forward sublayers maintain numerical stability and enable implicit 3-D coupling without heavy 3-D operators. A final $1 \times 1$ projection maps embeddings back to the forecast tensor with $T_{\text{out}}$ steps. The forward path consumes the node features and the per-level edge lists (horizontal and vertical) and returns $[B, T_{\text{out}}, C, N]$.

The unified core is trained by minimising the MSE, averaged across all forecast channels and spatial nodes

$$\mathcal{L}_{core} = \|\hat{X} - X\|_2^2.$$

This encourages accurate reconstruction of the full multivariate atmospheric state while keeping the representation compact and physically consistent.

**Task-specific heads and losses**

Two lightweight task heads extend the unified core to specialise in wind and precipitation forecasting.

**Wind head (u10/v10).** For 10-m wind, we attach a specialised two-channel head on top of the unified core that reads the shared embeddings and outputs $\hat{u}_{10}, \hat{v}_{10}$. Training uses a composite loss that augments component-wise MSE with a magnitude term and a directional cosine term:

$$\mathcal{L}_{wind} = \|(\hat{u}, \hat{v}) - (u, v)\|_2^2 + (\|\hat{v}\| - \|v\|)^2 + [1 - \cos(\hat{v}, v)],$$

where $\|(\hat{u}, \hat{v}) - (u, v)\|_2^2$ is a component MSE, $(\|\hat{v}\| - \|v\|)^2$ denotes magnitude penalty, and $1 - \cos(\hat{v}, v)$ is a direction penalty.

The additional terms encourage accurate wind strength and orientation, particularly under strong-wind conditions, while keeping the head computationally lightweight.

**Precipitation head (tp_log).** For precipitation, we use the same core and a single-channel head that predicts the log-transformed rate $tp_{log} = \log(1 + tp)$. The loss is a weighted MSE:

$$\mathcal{L}_{precip} = \sum w_{wet}(\hat{tp}_{log} - tp_{log})^2,$$

where weights $w_{wet}$ up-weight rainy nodes to counter the strong dry-pixel imbalance and stabilise learning on heavy-tailed rainfall distributions.

All components use the same base loss formulation defined in the forecasting objective, with task-specific adaptations of the general term $\ell(\hat{X}, X)$.

**Cross-scale forecasting and autoregressive rollout**

The model is trained in a one-step forecasting regime (+6 h lead) with an input history of with $T_{\text{in}} = 2$ (12 h) and a single output step $T_{\text{out}} = 1$. The unified core first learns to predict all meteorological channels jointly, after which the wind and precipitation heads are fine-tuned under the same regime. Training batches consist of tensors $[B, T_t, C, N]$ paired with static features and graph edges provided by the dataloader.

During evaluation, one-step models are rolled out autoregressively to generate multi-step forecasts up to +24 h (K = 4).

At each step, the predicted state becomes input for the next forecast. For specialised models, all channels use the general core's prediction except the target variables, which are overwritten by the respective head outputs (e.g., $u_{10}, v_{10}$ for wind, $\hat{tp}_{log}$ for precipitation). This preserves cross-variable coupling from the unified core while allowing each head to refine its own target field across the rollout horizon.

## Results and evaluation

We evaluate our MR-GNF against uniform-grid and single-scale baselines, focusing on (i) short-range skill, (ii) stability under autoregressive (AR) rollout, and (iii) efficiency under tight compute budgets.

All models were trained on a single RTX 6000 Ada GPU. The general backbone (1,625,545 params) converged after 12 epochs (~3.5 h per epoch, ≈42 GPU-hours), followed by task-specific 3-epoch fine-tunes for *wind* and *precipitation* (≈6 h per epoch, ≈18 GPU-hours each per head). The total cost of the full study was ~78 GPU-hours within single-device capacity.

Despite this compact setup, the model reaches strong accuracy for 0.25° forecasts across four key variables.

**2-m temperature (t2m)**

The general model keeps +6 h MAE near 0.46 K and exhibits nearly linear growth to +24 h (Fig. 3), demonstrating numerical stability under rollout.

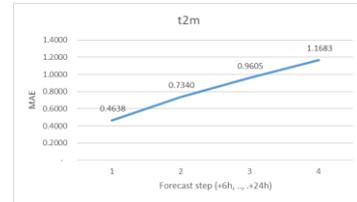

Figure 3: 2-m temperature MAE (+6…+24 h)

Spatial gradients remain realistic, though slightly diffused, reflecting the low-pass character of the GNN convolution on the irregular mesh.

**10-m wind (u10/v10)**

Wind components exhibit a similar pattern but benefit clearly from the task-specific head. At +6 h, the specialised model reduces MAE by 19-22 % (u10: 0.60 vs 0.75; v10: 0.63 vs 0.78) and maintains double-digit RMSE gains through +12 h (Fig. 4).

By +24 h, the head retains smaller errors than the general core, confirming that replacing $(\hat{u}_{10}, \hat{v}_{10})$ into the AR loop stabilises directional coherence and amplitude across steps. By +24 h, both heads retain lower errors than the general core, demonstrating consistent cross-scale generalization.

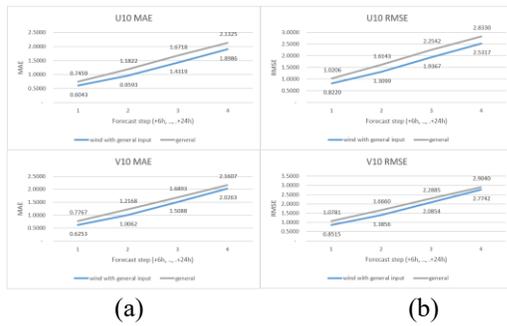

Figure 4: (a) U10/V10 MAE and (b) U10/V10 RMSE for the general model and the specialised wind head.

**Precipitation (tp_log)**

Using weighted loss on the log-transformed precipitation field enhances sensitivity to wet regions and reduces MAE by ≈ 12-18 % and RMSE by ≈ 2-5 % relative to the general model (Fig. 5).

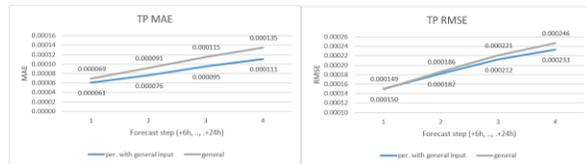

Figure 5: Total precipitation MAE/RMSE for the general and precipitation-specialised models.

With ~18 GPU-hours per head, the precipitation model improves MAE by ~12-18% and RMSE by ~2-5% at +12…+24 h; the wind model improves RMSE by ~10-21% (largest at +6…+12 h).

**Takeaways vs. Budget**

A ~1.63M-parameter backbone trained in ~42 GPU-hours delivers competitive t2m and baseline tp/wind skill at 0.25° ROI. All results are obtained without global cascades or multi-GPU runs, underscoring the accuracy-per-compute efficiency of the multi-scale mesh and spatio-vertical GNN design for regional short-range forecasting.

# Case study: Storm Henk over the British Isles (2 January 2024, UTC)

To examine real-world behaviour, we evaluate the model on Storm Henk, a fast-moving extratropical cyclone that swept across England and Wales on 2 January 2024, bringing intense rainfall and strong winds. This event aligns perfectly with the 6-hour model cadence and traverses the ROI from southwest to east, simultaneously testing all four surface prediction targets: temperature (t2m), precipitation (tp_log), and wind components (u10, v10).

**Event timeline (UTC)**

- 2024-01-02 00:00 – Approach of the frontal system; relatively clean pre-event background prior to intense rain.
- 2024-01-02 06:00 – Rain intensifies over the western ROI; near-surface wind increases ahead of the front.
- 2024-01-02 12:00 – Event peak across central/southern Great Britain; radar and analyses indicate widespread rain—ideal for evaluating tp_log and the wind vectors.
- 2024-01-02 18:00 – System exits eastward; rainfall tapers with a windy post-frontal wake.

**Surface wind component (u10/v10)**

Figure 7 compares ground truth (GT) and predicted near-surface wind components (u10 and v10) at +6 h, +12 h, +18 h, and +24 h leads.

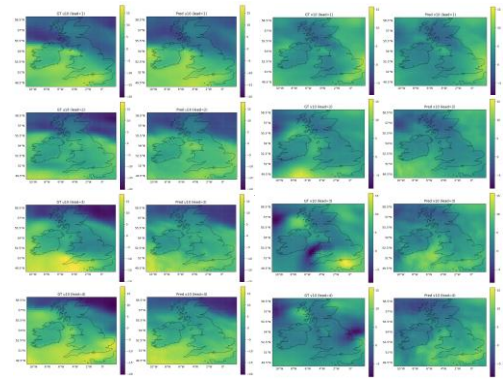

Figure 6: Modelled and observed u10/v10 wind components during Storm Henk at +6 h, +12 h, +18 h, and +24 h leads. Left: ground truth (ERA5); right: MR-GNF predictions.

At +6 h (lead 1), the model correctly reconstructs the dominant west-to-east flow over southern England and the coastal wind maxima, though localized jets appear slightly smoothed.

By +12 h (lead 2), the overall structure and orientation remain realistic, but mild amplitude loss emerges along the western approaches.

At +18 h (lead 3), deviations become most visible in *v10*: observed southerly cores over central England are reproduced but weakened and spatially spread.

At +24 h (lead 4), broad flow patterns persist while peak magnitudes are further damped, particularly in v10 along the system's eastern exit.

**Error profile and interpretation.** A small negative bias appears on speed peaks, consistent with the low-pass smoothing inherent to the GNN convolution. Spatial displacement errors are minimal; most deviations stem from amplitude damping rather than positional drift. A fast west-to-east frontal passage sets tight gradients and wind maxima near the Channel/Irish Sea. The model follows the advective evolution well, but amplitude loss grows with AR rollout (leads 3-4), consistent with compounding 1-step smoothing at each lead, trading short-term stability for amplitude fidelity.

**Near-surface temperature (t2m)**

Figure 7 (a) presents the 2 m temperature fields from the model and ERA5 ground truth.

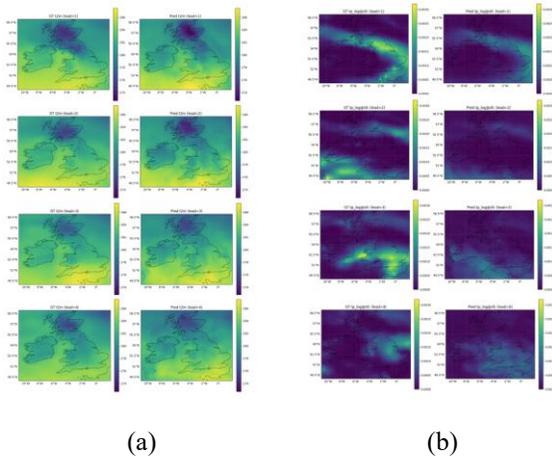

(a) (b)

Figure 7: Modelled and observed (a) 2 m temperature fields and (b) log-transformed total precipitation (tp_log) at +6 h, +12 h, +18 h, and +24 h during Storm Henk. Left: ground truth (ERA5); right: MR-GNF predictions.

At +6 h (lead 1), the model reproduces the north-south thermal contrast, with a well-placed warm sector over southern England and a cold pool across Scotland. The latter appears slightly too warm, reflecting early gradient damping.

By +12 h (lead 2), the front remains well positioned, though gradients are more diffuse, with a mild warm bias in the cold sector and a small under-shoot of the warmest values in the south—net effect is a reduced frontal contrast.

At +18 h (lead 3), the frontal zone softens further: the cold tongue over northeast England is under-represented, and southern warmth is muted by roughly 1 K (visually).

At +24 h (lead 4), large-scale structure persists, but extremes are most damped. The warm sector along SE England is underpredicted; the northern cold pool remains too warm, yielding the smallest gradient of all four leads.

**Error profile and interpretation.** Temperature extremes converge toward the mean, with a mild warm bias in cold regions and cool bias in warm ones. Displacement errors remain negligible. Storm Henk's strong baroclinic gradients expose the model's diffusive tendency under autoregressive operation. The MR-GNF tracks the frontal position accurately but gradually damps diurnal extremes, suggesting a characteristic low-pass filter response similar to that seen in wind and precipitation fields.

**Precipitation (tp_log)**

Figure 7 (b) shows the log-scaled total precipitation (tp_log) from ERA5 and model outputs across four leads.

At +6 h (lead 1), the model identifies the main southwest northeast rainband but underestimates its intensity; peak cores are clipped, and drizzle dominates.

At +12 h (lead 2), observed maxima shift southwest toward the Channel, whereas predictions lag and weaken, leading to several false negatives over the western approaches.

At +18 h (lead 3), the rainband re-forms over the southern Midlands; the model captures its placement but remains too smooth to reproduce mesoscale maxima.

At +24 h (lead 4), residual rainfall persists over southeast England; the model follows the track but keeps intensities too light and spatially fragmented.

**Error profile and interpretation.** Amplitude underestimation dominates all leads, with minimal displacement. The field exhibits clear low-pass smoothing and regression-to-mean effects from the log-scaled target formulation. While the model successfully tracks the temporal and spatial evolution of the rainband but systematically attenuates heavy-rain cores. This behaviour aligns with the compressive nature of the tp_log regression and with cumulative smoothing under autoregressive rollout.

**Summary and insight**

Across variables, Storm Henk illustrates the model's capacity to preserve spatial coherence and temporal continuity over a full day of evolution, despite its lightweight single-GPU setup. Errors are dominated by amplitude damping rather than displacement, confirming that the multi-resolution GNN effectively maintains phase alignment and cross-variable consistency while gradually diffusing extremes – a trade-off typical for stability-prioritised autoregressive systems. These qualitative patterns reinforce the quantitative findings: compact graph models can deliver credible 0.25° forecasts of complex mid-latitude storms while remaining computationally accessible.

## Limitations & Future Work

While MR-GNF demonstrates strong stability and cross-scale skill, several limitations remain. First, amplitude damping of extreme values, especially in wind gusts and precipitation, reflects the smoothing bias of graph convolutions and autoregressive rollouts. Future work should explore temporal diffusion correction or spectral sharpening losses to restore intensity without compromising stability. Second, the model currently relies on ERA5 reanalysis, which provides perfect boundary conditions. We recognise that applying it to real-time data streams will require bias-aware normalisation and assimilation of noisy inputs. Third, the study is limited to a single regional configuration (UK–Ireland sector); extending the tri-band mesh to diverse climatic regions would test generalisation and domain transfer. Finally, incorporating probabilistic or ensemble formulations, e.g., diffusion-based uncertainty quantification, could improve representation of storm predictability and yield actionable confidence intervals for operational forecasting.

## Conclusion

We presented MR-GNF, a lightweight regional weather-forecasting framework that combines a multi-scale ellipsoidal mesh, a unified spatio-vertical GNN, and targeted fine-tuning for wind and precipitation. Trained on ERA5 data for 1980-2024, the model achieves stable 6-24 h forecasts over Great Britain with MAE ≈ 0.46 K ($t_2$m), ≈ 0.6 m s$^{-1}$ ($u_{10}$/$v_{10}$), and log-precipitation RMSE ≈ $1.5 \times 10^{-4}$, while requiring only a single RTX 6000 Ada GPU and < 4 h per epoch. Case analysis of Storm Henk (Jan 2024) shows that the system reproduces the large-scale advection and baroclinic gradients with realistic timing, though peak intensities remain smoothed. The results demonstrate that geometry-aware, cross-scale GNNs can deliver credible regional forecasts on standard hardware, opening a path toward fast, sustainable, and operationally deployable AI-driven weather systems.